\documentclass{article}

\usepackage[nonatbib,preprint]{neurips_2022}

\usepackage[utf8]{inputenc} % allow utf-8 input
\usepackage[T1]{fontenc}    % use 8-bit T1 fonts
\usepackage{hyperref}       % hyperlinks
\usepackage{url}            % simple URL typesetting
\usepackage{booktabs}       % professional-quality tables
\usepackage{amsfonts}       % blackboard math symbols
\usepackage{nicefrac}       % compact symbols for 1/2, etc.
\usepackage{microtype}      % microtypography
\usepackage{xcolor}         % colors
\usepackage{enumitem}
\usepackage{comment}

\usepackage{subcaption}

\usepackage{graphicx}

\newcommand{\newcite}{\cite}

\title{Cultural Re-contextualization of Fairness Research \\in Language Technologies in India}

\author{
Shaily Bhatt \\ Google Research \\ shailybhatt@google.com \And
Sunipa Dev \\ Google Research \\ sunipadev@google.com \And
Partha Talukdar \\ Google Research \\ partha@google.com \AND
Shachi Dave* \\ Google Research \\ shachi@google.com \And
Vinodkumar Prabhakaran* \\ Google Research \\ vinodkpg@google.com}

\begin{document}
\maketitle

\begin{abstract}
Recent research has revealed undesirable biases in NLP data and models. However, these efforts largely focus on social disparities in the West, and are not directly portable to other geo-cultural contexts. In this position paper, we outline a holistic research agenda to re-contextualize NLP fairness research for the Indian context, accounting for Indian \textit{societal context}, bridging \textit{technological} gaps in capability and resources, and adapting to Indian cultural \textit{values}. We also summarize findings from an empirical study on various social biases along different axes of disparities relevant to India, demonstrating their prevalence in corpora and models.
\end{abstract}

\section{Introduction}
\label{sec1_intro}

Recent research has demonstrated that language technologies may capture, propagate, and amplify societal biases \cite{blodgett2020language}. While Natural Language Processing (NLP) has seen global adoption, most studies on assessing and mitigating such biases are situated in the Western context,\footnote{We use \textit{Western} or \textit{the West} to refer to the regions, nations and states consisting of Europe, the U.S., Canada, and Australasia, and their shared norms, values, customs, religious beliefs, and political systems \cite{kurth2003western}.} focusing primarily on axes of disparities prevalent in the Western public discourse, and hence not directly portable to non-Western contexts \cite{sambasivan2021re}. This is especially troubling in the case of India, a pluralistic nation of 1.4 billion people, with fast-growing investments in NLP research, development, and deployments from the government, the industry, and the startup ecosystem.
% \footnote{In government (\url{bhashini.gov.in)} and private sector (\url{tinyurl.com/indiaai-top-nlp-startups}, \url{tinyurl.com/google-idf-language}).}
While there is some recent work on NLP fairness in Indian languages like Hindi, Bengali, and Telugu \cite{pujari2019debiasing, malik2021socially}, re-contextualizing NLP fairness for the Indian context requires a holistic approach that accounts for the various relevant axes of social disparities in the Indian society, their proxies in language data, the disparate NLP capabilities across Indian languages and dialects, and the (lack of) availability of resources that enable fairness evaluations and mitigation \cite{sambasivan2021re}. 
In this paper, we summarize takeaways from an empirical analysis of biases in NLP models along various axes of disparities relevant in the Indian context, and then propose a holistic roadmap for re-contextualizing data and model fairness in NLP.

\vspace{-5pt}
\section{Summary of Empirical Results}

We first report some highlights from our extensive empirical analysis of social biases in NLP models in the Indian context \cite{bhatt2022re}. The axes of disparities we consider include two India-specific axes: a) \textit{Caste}, which is an inherited hierarchical social identity, that has been the basis of historical marginalization; and b) \textit{Region}, or ethnicity associated with geographic regions of India, as well as four globally-salient axes that have unique manifestations in the Indian context: a) \textit{Gender}, where there are different structural disparities in engagement of women in society as compared to the West; b) \textit{Religion}, wherein the majority and minority religious groups differ compared to the west; c) (dis)\textit{Ability} and 4) \textit{Gender Identity and Sexual Orientation}, around which the social discourse and awareness in India is fairly recent.
% We study biases along the axes of Caste, Region, Religion and Gender. 
We analyzed various proxies in language data for these social groups such as identity terms, personal names, and dialectal features to study biases in NLP models.

\begin{figure*}[t]
\centering
    \begin{subfigure}[t]{0.28\textwidth}
    \includegraphics[width=\linewidth]{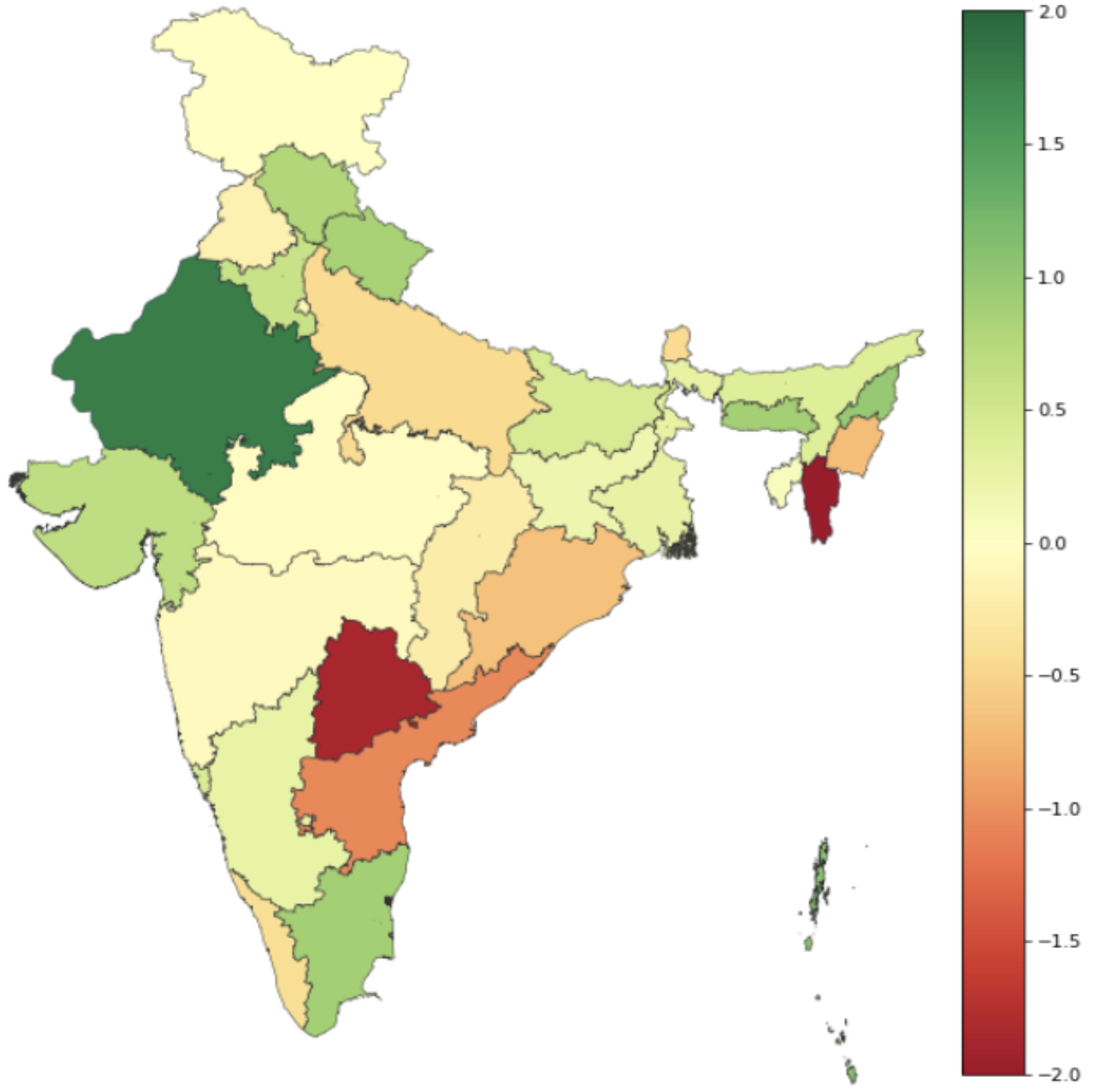}
    \caption{Region}
    \end{subfigure}\hspace{5pt}
    \begin{subfigure}[t]{.36\textwidth}
    \includegraphics[width=\linewidth]{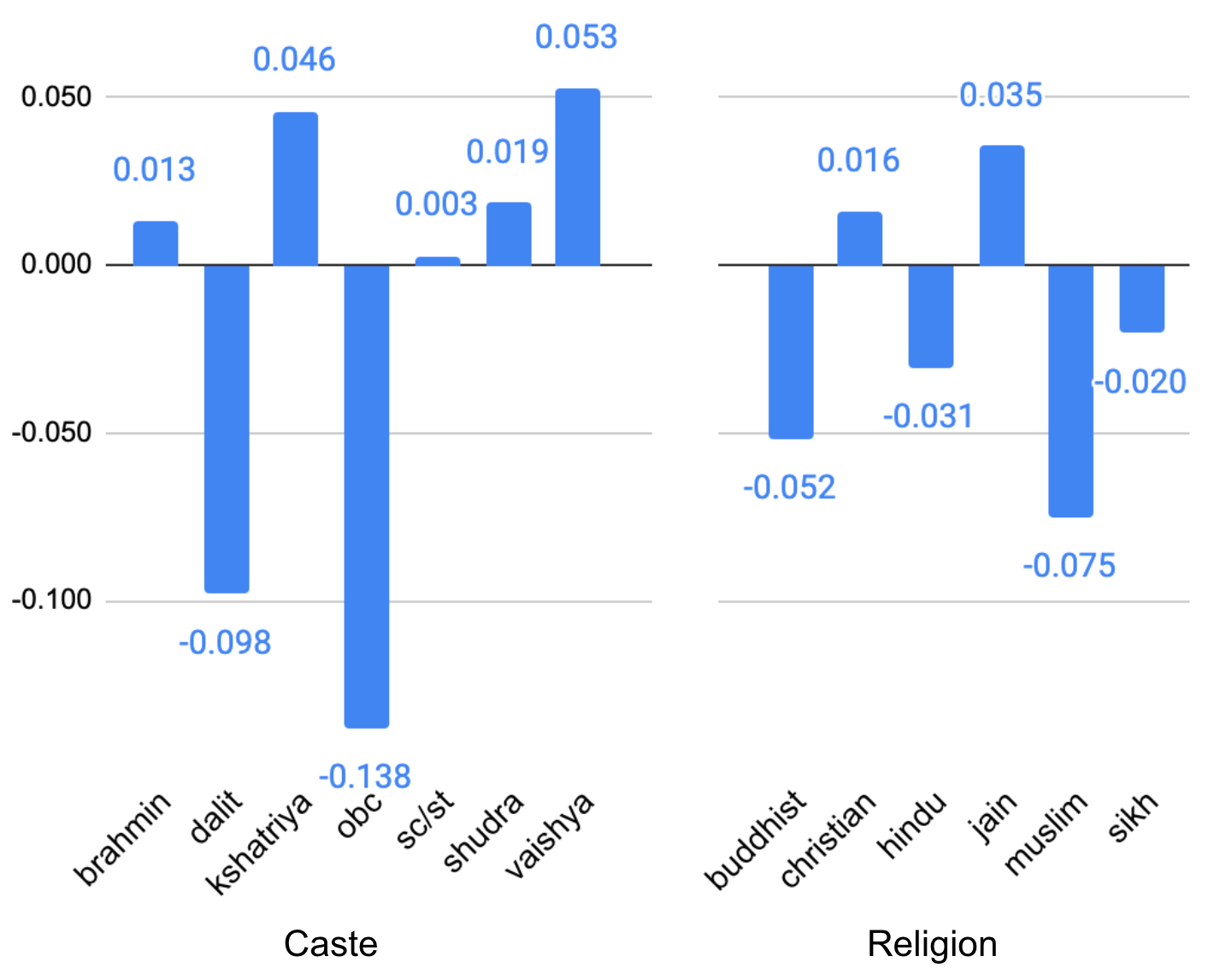}
    \caption{Caste and Religion}
    \end{subfigure}\hspace{5pt}
    \begin{subfigure}[t]{0.3\textwidth}
    \includegraphics[width=\linewidth]{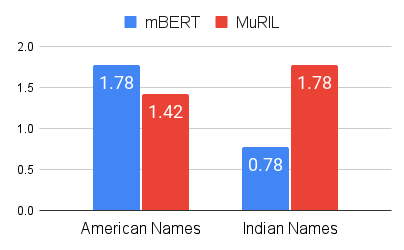}
    \caption{Gender}
    \end{subfigure}
\caption{Highlights from an empirical analysis of biases along axes of disparities in the Indian context. Fig (a) and (b) show perturbation analysis results \cite{prabhakaran2019perturbation} using identity terms for \textit{Region}, \textit{Caste}, and \textit{Religion} on the HuggingFace default sentiment model. Fig (c) shows the DisCo metrics \cite{filbert-paper} using Indian and American names to measure \textit{Gender} bias in language models mBERT and MuRIL.}
\label{fig_results}
\end{figure*}

% Figure~\ref{fig_results} shows some high-level results. 
Figures~\ref{fig_results}a and ~\ref{fig_results}b shows shifts in sentiment scores in response to perturbation analysis \cite{prabhakaran2019perturbation} of identity terms for \textit{Region}, \textit{Caste}, and \textit{Religion} on the HuggingFace default sentiment model (a DistillBERT fine-tuned with SST-2). In particular, we see that the model has learnt to associate higher negative sentiment towards marginalized sub-groups, such as `Dalit' and `OBC' (other backward castes) in caste, and `Muslim' in religion. For state identities, the model has learnt to associate more negative sentiment with southern states like Andhra Pradesh and Telangana, and North-Eastern states like Mizoram and Manipur. Figure~\ref{fig_results}c shows that DisCo metric \cite{filbert-paper} that measures whether the predictions of a model have statistically significant associations to (binary) gender in language models require Indian names with gender association in order to correctly detect encoded biases. 
In addition, we also built a human-curated dataset of stereotypes around Region and Religion axes to demonstrate that such stereotypes are preferentially encoded in models and corpora (not shown in the figures above). 

\vspace{-5pt}
\section{Towards Cultural Re-contextualization of NLP Fairness in India}
\label{sec2_background}

\begin{figure*}[t]
\centering
\includegraphics[width=\linewidth]{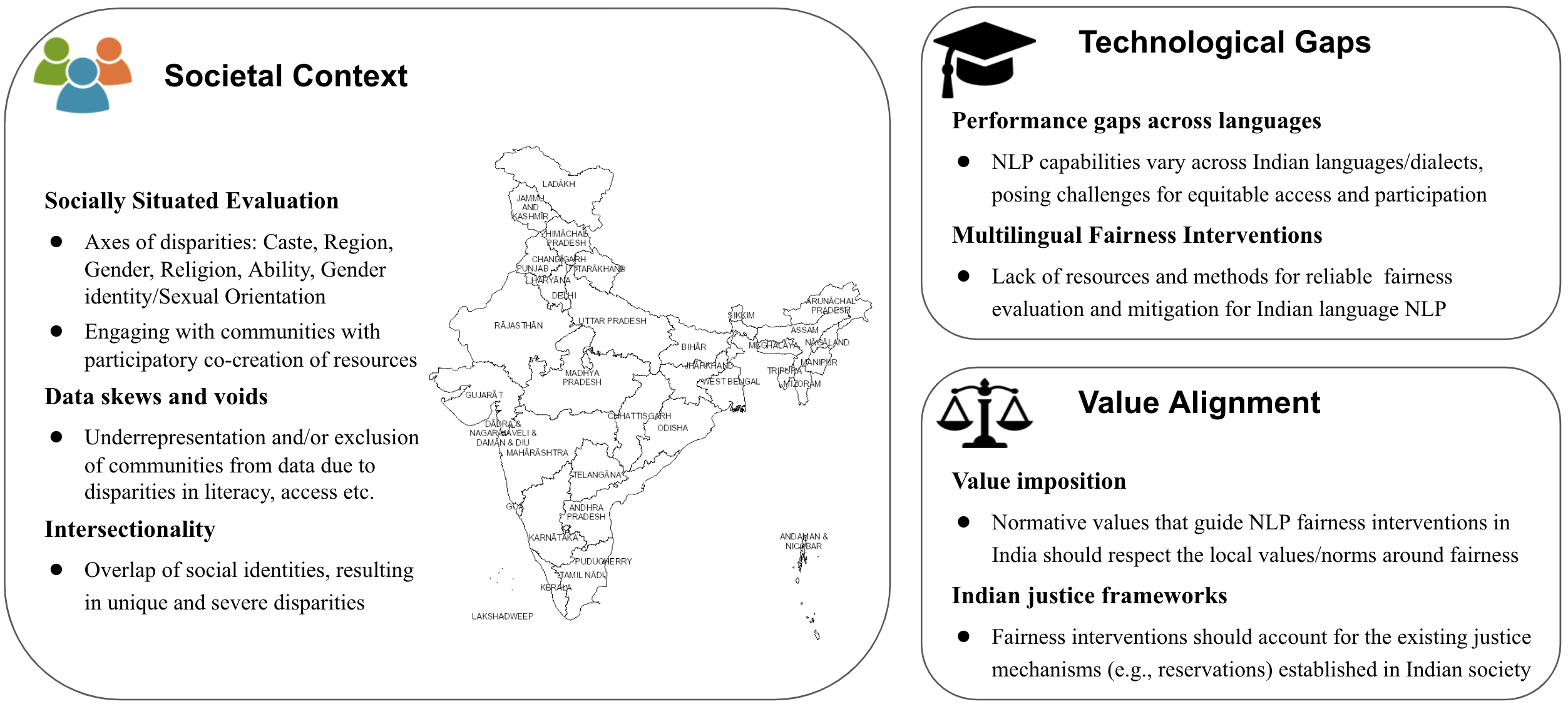}
\caption{A holistic research agenda for NLP Fairness in the Indian context: accounting for societal disparities in India, bridging technological gaps in NLP capabilities/resources, and adapting fairness interventions to align with local values and norms. (Map: \url{https://indiamaps.gov.in/soiapp/})
% \caption{Social disparities and proxies in language relevant for NLP Fairness in India. (Map source: \url{https://indiamaps.gov.in/soiapp/})
% NLP Fairness in the Indian Context. \vp{Only a placeholder figure; Feel free to design your own at \url{go/bindi-paper-figures-deck}; Objective: a pictorial representation of the agenda} \vp{Map source: \url{https://worldmapwithcountries.net/2020/03/12/india-map-with-states} to confirm copyright}
}
\label{fig:Indiaframework}
\end{figure*}

The above results demonstrate that NLP models reflect societal biases around socio-demographic subgroups in the Indian context. To effectively address these issues we need a holistic perspective that accounts for the various factors in the ecosystem. Building on \cite{sambasivan2021re}, we propose a holistic research agenda (Figure~\ref{fig:Indiaframework}) for re-contextualizing fairness in NLP along three dimensions: accounting for the \textit{societal context}, bridging the \textit{technological gaps}, and adapting to the local \textit{values and norms}.

\vspace{-3pt}
\subsection{Accounting for Indian \textit{Societal} context}

% \vspace{-5pt}
\paragraph{Socially Situated Evaluation:}
A major hurdle in accounting for different axes is the access to diverse annotator pools who have familiarity and lived experiences of the marginalized groups. This is important for fairness work in general \cite{denton2021whose}, but especially in India where public discourse around (dis)ability, gender identity and sexual orientation is relatively limited. 
Participatory approaches \cite{lee2020human} to co-create resources for fairness evaluation will be crucial for meaningfully addressing this gap.
% We believe that participatory approaches such as the Masakhane effort are crucial for addressing this gap in Fairness research.\footnote{\url{https://www.masakhane.io/}}

\vspace{-5pt}
\paragraph{Data Voids:}
Entire communities may be excluded from language data due to disparities in literacy and internet access \cite{sambasivan2021re}. 
% Although recent efforts have initiated building language models using data sourced from India \cite{khanuja2021muril}, 
Not accounting for such data voids might result in biases being baked into the language models that has become base infrastructure for NLP \cite{bommasani2021opportunities}.   
Further, the risk of unintentionally excluding marginalized communities based on dialect or other linguistic features while filtering data to ensure quality \cite{dodge2021documenting,gururangan2022whose} is even higher in the Indian context because of very limited computational representation of marginalized communities. 
Participatory data curation (e.g., collecting language data specifically from marginalized communities \cite{abraham-etal-2020-crowdsourcing, nekoto-etal-2020-participatory} can significantly help bridge such data voids.
% Participatory approaches to data curation such as that done by \citet{abraham-etal-2020-crowdsourcing} to collect speech data from tribal and rural Maharashtra can significantly help bridge such data voids.

% While large language models have become standard infrastructure for NLP research and development \cite{bommasani2021opportunities,bender2021dangers}, they are largely trained on data sourced from the West. For instance, \citet{johnson2022ghost} point out that popular language models such as GPT-3 exhibit dominant US values, when analyzed with value sensitive topics. While recent efforts have initiated work on building large language models with data from Indian contexts \cite{khanuja2021muril}, these are prone to misrepresentation and exclusion of entire communities \cite{sambasivan2019toward}, who are typically already marginalized. Such data typically come from internet sources like social media, Wikipedia, books, and news corpora \cite{kakwani2020indicnlpsuite, khanuja2021muril}; however, internet participation is limited to those with education, access to technology, and digital awareness. Further, the risk of unintentionally excluding marginalized communities based on dialect or other linguistic features while filtering data to ensure quality \cite{dodge2021documenting,gururangan2022whose} is even higher in the Indian context because of very limited computational representation of marginalized communities. Thus, data voids should be taken into consideration while building data resources and models.

\vspace{-5pt}
\paragraph{Intersectionality:}
% The overlap of various social identities, such as religion, gender, sexuality, and class, contributes to the specific type of systemic oppression and discrimination experienced by an individuals or subgroups \cite{collins2020intersectionality}.
Due to the interplay of all the diverse axes in the Indian context, intersectional biases 
% \cite{collins2020intersectionality} 
experienced by different marginalized groups are further exacerbated \cite{dalit-women-in-india}. With notable differences in literacy, economic stability, technology access, and healthcare access across geographical, caste, religious, and gender divides, representation in and access to language technologies is also disparate. Bias evaluation and mitigation interventions should account for these intersectional biases.
% by marginalised communities.

% \paragraph{Participatory Design}

% The purpose of the dataset we create is to serve as a first step towards evaluating bias in models and data in India specific context. However, having a limited number of annotators, this may not capture pluralistic nuances. Further, without involvement of the communities that face marginalization, there is a risk of not capturing their lived experiences. For example, in our dataset, one of the gender sub-groups is "transgender", however none of the annotators we have identify as transgender (to the best of our knowledge). This limits the capability of typical NLP style annotation practices (such as this one) to capture the lived experiences of marginalized communities. More so because data workers employed by annotation firms and crowd-sourcing platforms tend to be urban, educated, and digitally-literate \cite{abraham-etal-2020-crowdsourcing} who represent a elite fraction of the Indian society. To avoid creating datasets that have a myopic view point, especially in inherently subjective tasks like Stereotype annotations, participatory design with active involvement of communities on the ground is necessary to create reliable resources reflect the ground reality of marginalization.

\vspace{-3pt}
\subsection{Bridging cross-lingual \textit{Technological} gaps}

% While we focus on English language data and models in this paper, it is crucial to mitigate the gaps in NLP capabilities and resources across Indian languages, both in general and for fairness research. 

% \vspace{-5pt}
\paragraph{Performance gaps across languages:}
% \paragraph{Fair Performance across Languages:}
Although India is a vastly multilingual country with hundreds of languages, and thousands of dialects, there are wide disparities in NLP capabilities across these languages and dialects. These disparities hinder equitable access, creating barriers to internet participation, information access, and in turn, representation in data and models. While the Indian NLP community has made major strides in bridging this gap (e.g., \cite{khanuja2021muril}), more work is needed in building and improving NLP technologies for marginalized and endangered languages and dialects.

% Although India is a vastly multilingual country with 22 recognised languages, and thousands of varieties and dialects, there are wide disparities in NLP technology capabilities across these languages and dialects. While more work needs to be done to assess and mitigate effects of unfair societal biases in NLP in Indian languages, disparate performance across these languages pose fairness challenges regardless of those biases. Poor technology in low-resource languages can in turn create barriers to internet participation, information access, and cyclically, representation in data and models.  It is thus essential that the performance of language technologies is not widely disparate for various (marginalised) sub-groups preventing them the opportunity to access and utilize technology effectively

\vspace{-5pt}
\paragraph{Multilingual fairness research:}
NLP Fairness research relies on evaluation resources that are currently largely built in and for the Western context.
% While we present such resources recontextualized for the Indian context in this paper, we limited our focus to only English. 
It is crucial to build these resources in Indian languages, along the lines of recent work on Hindi, Bengali, and Telugu \cite{malik2021socially,pujari2019debiasing}, since biases may manifest differently in data and models for different languages, and how bias transfers in transfer-learning paradigms for multilingual NLP is unknown. Finally, bias mitigation in one (or a few) language(s) may have counter-productive effects on other languages. Hence, a research agenda for fair NLP in India should address these various unknowns that the multilingual setting brings.

\vspace{-3pt}
\subsection{Adapting to Indian \textit{Values and Norms}}

% Fairness interventions essentially impart a normative value system on model behaviour. It is crucial to ensure that these interventions are not at odd with Indian values, norms, and legal frameworks. 

% \vspace{-10pt}
\paragraph{Avoiding value imposition:}
Fairness inquiries answer questions such as: what does it mean to be fair or unfair, and how fair is fair enough? These questions, and their answers, are rarely made explicit; rather a shared understanding is implicitly assumed, risking value imposition. For instance, these answers often draw largely from Western values of fairness that are rooted in {egalitarianism, consequentialism, deontic justice, and Rawls’ distributive justice} \cite{sambasivan2021re}. However, the philosophy of fairness in India is rooted in social restorative justice. More work should look into such value alignment challenges, which is not trivial when it comes to deploying fairness interventions \cite{gabriel2020artificial,prabhakaran2022human}.

\vspace{-5pt}
\paragraph{Accounting for Indian justice models:}
India has established restorative
justice measures in various resource allocation contexts, colloquially known as the ``reservations'' \cite{ambedkar2014annihilation}, where historically marginalized communities (such as Dalits, other backward castes, Adivasis (tribals), and religious minorities) are afforded fixed quotas in educational institutes and government jobs to counter historical deprivation. NLP fairness research in these domains should consider how fairness interventions work in the context of such established measures.

\newpage

{\small
\bibliographystyle{plain}
\bibliography{neurips_2022}
}

\end{document}